\documentclass[letterpaper, 10 pt, conference]{ieeeconf}
\IEEEoverridecommandlockouts
\usepackage{mathptmx}
\usepackage{amsmath,amssymb}
\usepackage{graphicx}
\usepackage[table,dvipsnames]{xcolor}
\usepackage{makecell,booktabs,tabularx,array,multirow}
\usepackage{adjustbox}
\usepackage{pifont}
\usepackage{float}
\usepackage{algorithm}
\usepackage{algpseudocode}
\usepackage{placeins}
\usepackage{cite}
\usepackage{url}
\usepackage{subcaption}
\let\labelindent\relax
\usepackage{enumitem}

\definecolor{TACOAccent}{RGB}{91,80,158}
\definecolor{TACOShade}{RGB}{241,239,250}

\definecolor{stagegreen}{RGB}{46,139,87}

\newcommand{\papertitle}{%
  \mbox{TACO\hspace{0.12em}%
    \raisebox{-0.10em}{%
      \includegraphics[height=0.80em]{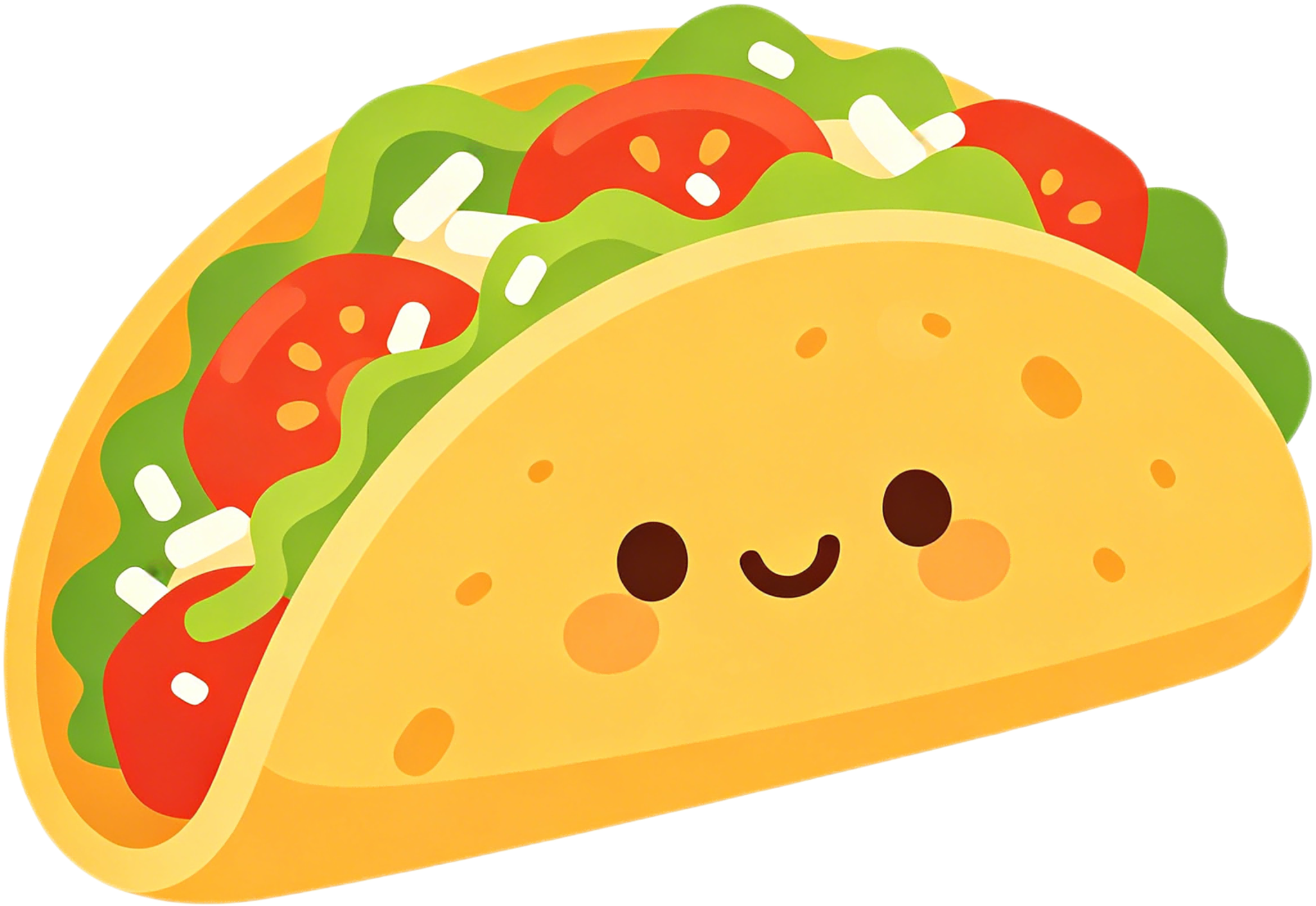}%
    }%
  }:
  TActile World Model as a Self-COrrector\\
  for Scalable Robot Policy Post-Training
}
\newcommand{\projecthomepage}{%
  \leavevmode\pdfstartlink attr{/Border [0 0 0]} user{%
    /Subtype /Link /A << /S /URI /URI (https://taco-wm.github.io/) >>}%
  \textcolor{blue}{\texttt{https://taco-wm.github.io/}}\pdfendlink}

\title{\LARGE \bf \papertitle}
\author{%
Shengbang Liu$^{1,3,*}$, Yueru Jia$^{1,2,*}$,
Yuyang Yan$^{1,*}$, Jiaming Liu$^{1,*,\dagger}$, Xinran Zhang$^{1,2,*}$\\
Qiuxuan Feng$^{1}$, Yandong Guo$^{2}$, Shiji Zhou$^{4}$,
Boxin Shi$^{1}$, Shanghang Zhang$^{1,\ddagger}$%
\thanks{$^{*}$Equal Contribution. $^{\dagger}$Project Lead. $^{\ddagger}$Corresponding Author.}%
\thanks{$^{1}$State Key Laboratory of Multimedia Information Processing, School of Computer Science, Peking University.}%
\thanks{$^{2}$AI$^2$ Robotics. $^{3}$Sun Yat-sen University. $^{4}$Beihang University.}%
\\[0.4em]
\normalsize\textbf{Project Page:} \projecthomepage
}

\begin{document}
\bstctlcite{TACO_bibcontrol}
\maketitle
\thispagestyle{empty}
\pagestyle{empty}

\begin{figure*}[t]
    \centering
    \includegraphics[width=\textwidth]{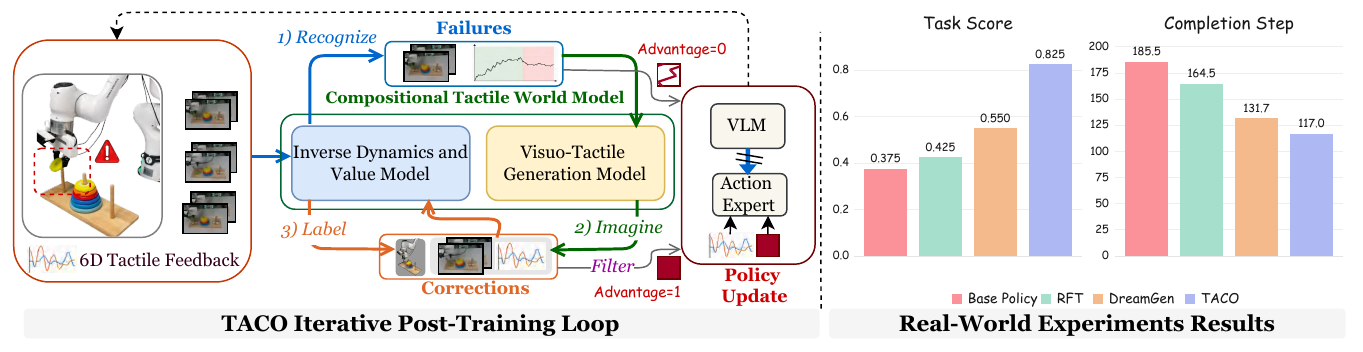}
    \caption{
        \textbf{Overview of TACO.}
        TACO uses a compositional tactile world model to turn real rollout
        failures into local visuo-tactile corrections for scalable robot policy
        post-training.
        Across six contact-rich real-world tasks, it improves average task
        score (TS) from 0.375 to 0.825 after two post-training iterations.
    }
   \label{fig:overview}
\end{figure*}

\begin{abstract}
Vision-Language-Action models and World Action Models have shown promising generalization in robotic manipulation but remain fragile in contact-rich tasks, where contact perturbations can cause failures that are difficult to detect from vision alone.
Corrective post-training with tactile feedback can improve recovery, but scaling such supervision through human intervention is costly.
World models can synthesize additional training data, yet vision-only generation may produce visually plausible but contact-inconsistent trajectories.
We therefore introduce \textbf{TACO}, a scalable robot policy post-training framework built on a compositional tactile world model.
Given real rollouts, TACO follows a \emph{Recognize--Imagine--Label} loop: an inverse dynamics and value model identifies failure-adjacent states using progress estimates, a visuo-tactile generation model imagines local corrections by jointly generating video and tactile sequences, and the inverse dynamics and value model labels them with corrective actions and progress scores.
Candidates are filtered for kinematic feasibility and tactile plausibility, then selected by predicted progress gain.
TACO aggregates demonstrations, real rollouts, and selected corrections for iterative post-training.
It combines \emph{knowledge-insulated tactile adaptation} with CFG-RL using binary advantage labels while keeping the pretrained VLM backbone fixed.
Experiments on real-world tasks show that TACO improves the average task score from 0.375 to 0.825 after two post-training iterations.
\end{abstract}
\section{Introduction}
\label{sec:intro}
Vision-Language-Action (VLA) models and World Action Models (WAMs) have shown promising progress in robotic manipulation by transferring vision-language priors to robot action generation~\cite{kim2024openvla,intelligence2025pi_,jia2025video2act}, but remain fragile in contact-rich tasks.
In real-world rollouts, failures often occur near contact transitions, where visual changes are subtle but tactile signals reflect slippage, insufficient pressure, or abnormal torque.
For example, in \textit{Wipe Whiteboard}, an eraser may cover a mark without sufficient pressure to remove it; in \textit{Twist Bottle Cap}, a gripper may align with the cap without generating effective twisting torque.
Such failures can occur even when the policy correctly identifies the task objective but cannot recover from unexpected contact changes.
This suggests that post-training with tactile corrective supervision can improve recovery from localized contact failures by focusing supervision on contact-sensitive stages rather than entire trajectories.

However, collecting corrective demonstrations is difficult to scale because it requires repeated human monitoring and manual recovery at failure states~\cite{ross2011reduction,hoque2024intervengen,xu2026compliant,wang2026interactive,li2026hi,xu2026twinrl}.
Recent works therefore explore world models to synthesize imagined rollouts for policy improvement~\cite{guo2026vlaw,zhu2025wmpo,yang2026rise,jiang2026wovr,jang2025dreamgen,yu2026wm}.
Yet vision-only world models may generate visually plausible rollouts with inconsistent contact dynamics.
This calls for joint visuo-tactile prediction and action inference to generate local corrective supervision.

Motivated by these observations, we propose \textbf{TACO}, a framework for scalable robot policy post-training built on a \emph{compositional tactile world model}.
Given real robot rollouts, TACO follows a \textit{Recognize--Imagine--Label} procedure.
First, an inverse dynamics and value model \emph{recognizes} failure-adjacent states by estimating signed progress from visual and tactile inputs.
Then a visuo-tactile generation model \emph{imagines} correction segments by jointly denoising future video and tactile sequences, with temporal RoPE aligning video and force tokens within self-attention.
Finally, the inverse dynamics and value model \emph{labels} these segments with corrective actions and progress scores.
Candidates are filtered for kinematic feasibility and tactile plausibility, then selected by predicted progress gain.

To incorporate tactile corrective supervision into robot policy post-training, a key challenge is that naively fine-tuning the full model may erode pretrained visual-language priors.
TACO therefore adopts \emph{knowledge-insulated tactile adaptation}, which freezes the pretrained VLM backbone and restricts updates to the tactile encoder, adaptation layers, and action expert~\cite{driess2026knowledge}.
To learn from both successful behavior and failed execution, TACO further adopts \emph{CFG-RL}~\cite{frans2025diffusion,intelligence2025pi}, which conditions action generation on binary advantage labels.
At inference, classifier-free guidance favors the positive-label action distribution.
Demonstrations, real rollouts, and selected imagined corrections are iteratively aggregated for post-training.

We evaluate TACO on six real-world manipulation tasks. After two post-training iterations, TACO improves the average task score from 0.375 to 0.825.
Ablation studies further validate the effectiveness of the imagined correction data and key framework components.
In summary, our main contributions are as follows:
\begin{enumerate}[
    label=\arabic*),
    leftmargin=*,
    topsep=2pt,
    partopsep=0pt,
    itemsep=2pt,
    parsep=0pt
]
    \item We develop a compositional tactile world model that combines joint denoising of video and force sequences with an inverse dynamics and value model for action inference and progress estimation.

    \item We propose an iterative \textit{Recognize--Imagine--Label} framework that generates local corrections from failure-adjacent states and selects them through validity filtering and progress-based ranking.

    \item We combine knowledge-insulated tactile adaptation with CFG-RL to learn from real experience and imagined corrections while keeping the pretrained VLM backbone fixed.
\end{enumerate}
\section{Related Work}
\noindent\textbf{World Models for Robot Learning.} 
World models predict future observations or visual states, and recent robot-learning methods use them to generate visual trajectories of task execution~\cite{feng2026harmowam}.
To turn imagined futures into policy-improvement signals, inverse dynamics models infer actions from generated videos~\cite{tan2025anypos,mi2026tc}, while reward models provide dense feedback for failure localization and rollout selection~\cite{tan2025robo,xu2026robodopamine2,liang2026robometer,zhai2025vlac,liu2026steam}.
Recent works use world models as simulators for RL-based policy refinement and policy post-training~\cite{jiang2025world4rl,xiao2025world,sharma2026world,jiang2026wovr,zhu2025wmpo,guo2026vlaw,liu2026world,yang2026rise,team2026gigabrain}. Together, these components and frameworks close the loop between imagined experience and policy improvement.
Unlike prior methods that rely on long-horizon prediction or online human correction~\cite{ross2011reduction,hoque2024intervengen,xu2026compliant,wang2026interactive,li2026hi,xu2026twinrl}, we use real rollouts to recognize failure states, imagine segments with tactile estimation, and relabel them into corrective action data for iterative robot policy post-training.

\noindent\textbf{Tactile Sensing for Robot Learning.}
Vision-Language-Action (VLA) models and World Action Models (WAMs) show strong semantic grounding and task generalization, but remain limited in contact-rich manipulation because physical interactions are weakly observable from vision alone.
Recent policies incorporate force--torque measurements or tactile images to improve contact awareness, force-sensitive control, and physical grounding~\cite{yu2026forcevla,huang2025tactile,zhang2026tacvla,ye2025dreamtacvla,zang2026tacforesight,ma2026tacpac}.
Beyond policy conditioning, visuo-tactile world models predict future contact dynamics, treating touch as part of environment dynamics rather than solely as an auxiliary policy input~\cite{zheng2026omnivta,higuera2026visuo,lou2026dreamtac,huang2026vitacworld,neoteai2026n0twam,xue2026hitacwam}.
However, naively adding tactile inputs can impair pre-contact perception and grounding~\cite{li2026vla}.
We use tactile feedback for both action prediction and force-conditioned world-model imagination.
Our knowledge-insulated training~\cite{driess2026knowledge} protects pretrained VLM representations from tactile-action losses, preserving VLA priors during corrective post-training.
\section{Method}
\begin{figure*}[t]
  \centering
  \includegraphics[width=\textwidth]{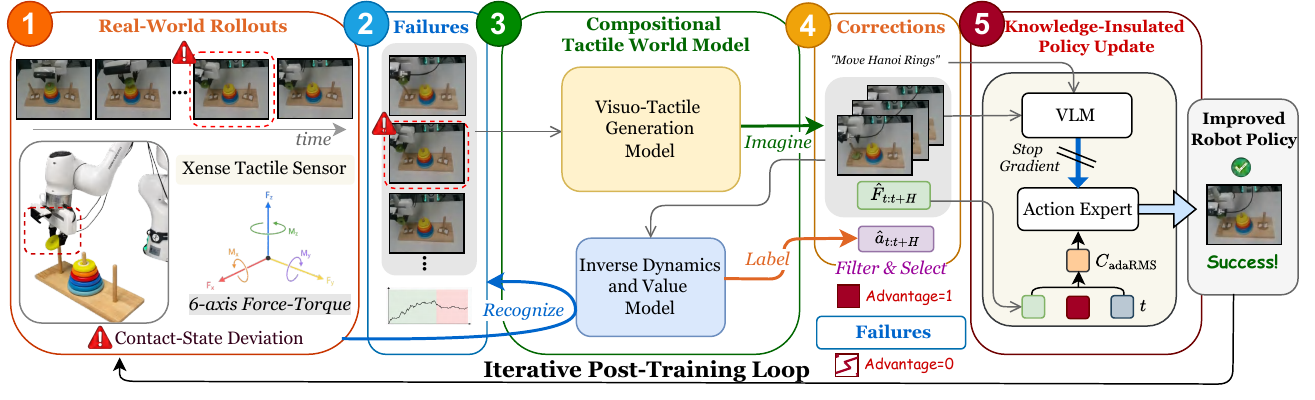}
    \caption{
        \textbf{TACO framework.}
        The compositional tactile world model follows a
        \textit{Recognize--Imagine--Label} loop to locate failure-adjacent
        states, generate visuo-tactile corrections, and infer actions and
        progress values.
        Corrections are filtered for validity, selected by predicted progress
        gain, and combined with demonstrations and real rollouts for CFG-RL
        post-training with knowledge-insulated tactile adaptation.
    }
  \label{fig:framework}
\end{figure*}

TACO builds on a \emph{Compositional Tactile World Model} to enable scalable robot policy post-training in contact-rich manipulation. 
As illustrated in Figure~\ref{fig:framework}, it converts policy failures in real-world rollouts into imagined correction data through a \emph{Recognize--Imagine--Label} loop: recognizing failure-adjacent states, imagining local correction segments, and labeling executable corrective actions.
The resulting correction data are iteratively aggregated with demonstrations and real rollouts for robot policy post-training.
% Sections~\ref{method:worldmodel} and~\ref{method:correction} introduce the compositional tactile world model and the iterative data aggregation procedure, respectively, while Section~\ref{method:training} details the post-training method.

\subsection{Compositional Tactile World Model}
\label{method:worldmodel}
\begin{figure*}[t]
  \centering
  \includegraphics[width=\textwidth]{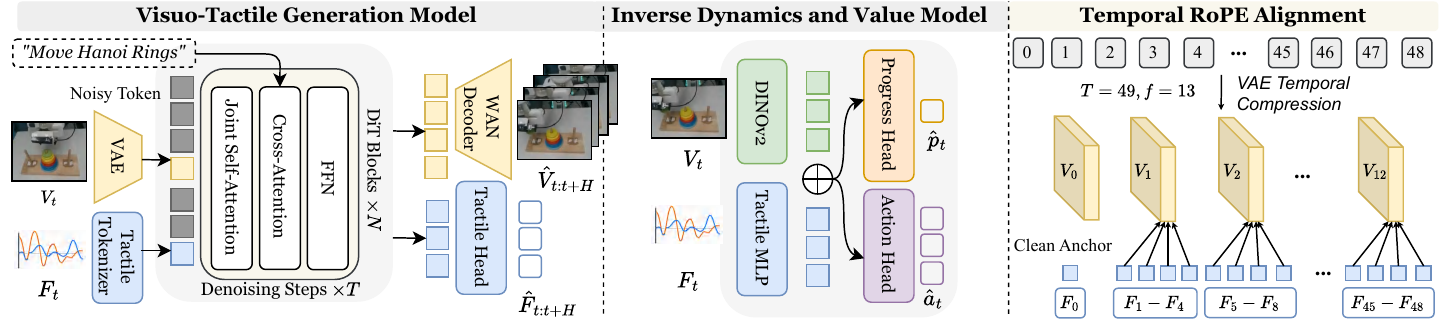}
    \caption{
        \textbf{Compositional Tactile World Model Architecture.}
        The visuo-tactile generation model synthesizes local corrections
        through joint denoising, using temporal RoPE to align tactile tokens
        with video latent tokens.
        The IDVM infers corrective actions and progress values from the
        generated visuo-tactile segments.
    }
  \label{fig:02}
\end{figure*}

As illustrated in Figure~\ref{fig:02}, the compositional tactile world model combines a visuo-tactile generation model for synthesizing local corrections from failure-adjacent states with an inverse dynamics and value model for action inference and progress estimation.

\noindent\textbf{Visuo-Tactile Generation Model.} 
We initialize the model from a Wan2.2-TI2V-5B~\cite{wan2025wan} backbone and perform additional pretraining on diverse robot data, including public datasets such as DROID~\cite{khazatsky2024droid}, AgiBot~\cite{bu2025agibot}, and RoboMIND~\cite{wu2024robomind}, together with closed-source robot data.
This additional pretraining provides priors for visual fidelity and robot--scene consistency.
We then fully fine-tune the model on sliding-window segments of our real-world demonstrations using flow matching, adapting it to contact-rich tasks through joint video--tactile denoising.

Given video latent tokens \(X^v\in \mathbb{R}^{B\times N_v\times d}\) and tactile sequence \(F\in\mathbb{R}^{B\times T\times 12}\), where \(B\), \(N_v\), \(T\), and \(d\) denote the batch size, number of video tokens, tactile sequence length, and hidden dimension, respectively, we tokenize tactile signals as \(X^f=T_\eta(F)\in \mathbb{R}^{B\times T\times d}\). 
Each tactile vector concatenates the left and right six-axis force--torque readings.
We concatenate video and force tokens as \(X=[X^v;X^f]\in \mathbb{R}^{B\times (N_v+T)\times d}\), enabling bidirectional video-force interaction within DiT self-attention. 
After denoising, the outputs are decoded into future video and tactile trajectories.
For training, video and force share the same sampled denoising timestep. 
Let \((\xi_1^v,\xi_1^f)\) denote clean video-latent and force segments, and \((\xi_0^v,\xi_0^f)\) denote the corresponding Gaussian noise. 
The model predicts video and force flow fields \(u_\psi^v\) and \(u_\psi^f\) with the joint flow-matching loss:
\[
\mathcal{L}_{\mathrm{joint}} = \left\|u_\psi^v-(\xi_1^v-\xi_0^v)\right\|_2^2 + \lambda_f\left\|u_\psi^f-(\xi_1^f-\xi_0^f)\right\|_2^2,
\]
where \(\lambda_f\) weights the tactile term.

We use temporal RoPE alignment and first-frame tactile anchoring to stabilize joint denoising. Since Wan2.2 applies RoPE over a 3D video latent grid, while tactile tokens only contain temporal force-torque information, we align each tactile token to the video latent temporal axis. 
Following the VAE temporal compression pattern, we align the
initial tactile token with latent position zero and each subsequent
group of four tactile tokens with the next latent temporal position.
For tactile sequence length $T$ and video latent temporal length
$f=1+(T-1)/4$, we assign
\[
\rho(i)=\left\lceil \frac{i}{4}\right\rceil,
\qquad i=0,\ldots,T-1.
\]
Tactile tokens use temporal RoPE at \(\rho(i)\), with spatial rotation factors fixed to the identity \(1+0j\).
We keep the initial tactile reading \(F_0\in\mathbb{R}^{12}\) unnoised as a first-frame anchor to reduce contact-state ambiguity and stabilize prediction.

\noindent\textbf{Inverse Dynamics and Value Model.}
Given an RGB frame \(I_t\) and its synchronized tactile signal \(F_t\in\mathbb{R}^{12}\), the model predicts an action and a signed progress score as \((\hat{a}_t,\hat{p}_t)=U_\phi(I_t,F_t)\), where $\hat a_t \in \mathbb{R}^{7}$ comprises three end-effector position components, three rotation angles, and a scalar gripper command, and $\hat p_t \in \mathbb{R}$ denotes the signed progress score.
The visual pathway combines DINOv2 with registers~\cite{oquab2023dinov2,darcet2023registers} and a direction-aware decoder for spatial grounding, while an MLP encodes tactile inputs.
The fused representation \(z_t=[z_t^v;z_t^f]\) feeds an action head \(\hat{a}_t=h_a(z_t)\) and a progress head \(\hat{p}_t=h_p(z_t)\), allowing contact cues to inform both predictions.
We train the model on successful and failed teleoperated trajectories spanning diverse contact conditions and stages of task execution.
We construct dense progress targets from manually annotated subtask keyframes.
Normalized progress is linearly interpolated between consecutive keyframes to obtain frame-level targets \(\bar{p}_t\in[0,1]\).
For failed trajectories, we negate the interpolated progress from failure onset onward.
Using signed progress as the value target, we jointly supervise action and progress predictions with
\[
\mathcal{L}_{\mathrm{IDV}}=\mathrm{SmoothL1}(\hat{a}_t,a_t)+(\hat{p}_t-p_t)^2.
\]
The trained model assigns corrective action and progress labels to imagined segments for robot policy post-training.

\subsection{Iterative Correction and Data Aggregation}
\label{method:correction}
\noindent\textbf{Recognize Failure-Adjacent States.}
At iteration \(k\), we deploy the current policy \(\pi_\theta^{(k)}\) to collect robot rollouts \(\mathcal{D}_{\mathrm{roll}}^{(k)}\).
For each rollout \(\tau\in\mathcal{D}_{\mathrm{roll}}^{(k)}\), the inverse dynamics and value model estimates a signed progress score \(\hat{p}_t\) at each timestep.
Anchor selection is restricted to failed rollouts, with terminal states masked out. We recognize failure-adjacent states where task progress stalls or decreases over a short temporal window:
\[
\mathcal{S}_{\mathrm{anchor}}^{(k)}=\{(\tau,t)\mid\hat{p}_{t+\Delta}-\hat{p}_t<\epsilon\},
\]
where \(\Delta\) is the window length and \(\epsilon\) is a progress-gain threshold.

\noindent\textbf{Imagine Visuo-Tactile Corrections.}
From each anchor, we sample $K_1=32$ correction segments
using independent denoising seeds. Each segment contains
$T$ frames including the anchor, with a future horizon
$H=T-1$. Conditioned on $(I_t,F_t,l)$, the visuo-tactile
generation model jointly denoises video and tactile sequences:
\[
(\hat I^{(j)}_{t:t+H},\hat F^{(j)}_{t:t+H})
\sim G_\psi(\cdot \mid I_t,F_t,l).
\]
Each candidate \(j\in\{1,\ldots,K_1\}\) captures local visual evolution and contact dynamics around the anchor state.

\noindent\textbf{Label Actions and Values.}
The inverse dynamics and value model labels each candidate frame by frame:
\[
(\hat{a}_s^{(j)},\hat{p}_s^{(j)})=U_\phi(\hat{I}_s^{(j)},\hat{F}_s^{(j)}),
\]
for \(s=t,\ldots,t+H\).
The predicted actions specify candidate robot commands for kinematic checks, while the progress estimates support value-based selection.

\noindent\textbf{Filter and Select Candidates.}
We first apply validity filtering based on kinematic feasibility and tactile plausibility.
For each candidate, we reconstruct the commanded trajectory from the recorded anchor configuration and check for a continuous inverse-kinematics solution satisfying robot-controller limits and workspace constraints.
We also check predicted force--torque magnitudes and temporal changes against task-specific bounds calibrated on valid real trajectories.
Valid candidates are ranked by predicted progress gain
\(g_j=\hat{p}_{t+H}^{(j)}-\hat{p}_t\), where \(\hat{p}_t\)
is the progress estimate at the real anchor state.
We retain up to \(K_2=8\) candidates with the highest gains satisfying \(g_j\geq\tau_p\).
Retained candidates receive the positive recovery label \(y=1\) and are added to the accumulated correction dataset for robot policy post-training with CFG-RL.
We assess physical executability separately through real-world replay of a sampled subset.

\subsection{Knowledge-Insulated Post-Training}
\label{method:training}

\noindent\textbf{Tactile Adaptation.}
To preserve pretrained visual-language knowledge for pre-contact perception and grounding, we freeze the VLM backbone and restrict post-training to the tactile encoder, adaptation layers, and action expert~\cite{driess2026knowledge}.
The VLM encodes image, language, and robot-state tokens into prefix representations \(z_t\).
Tactile history and the advantage label used by CFG-RL condition only the action expert through adaRMSNorm.

\noindent\textbf{CFG-RL Training.}
At each iteration, we apply CFG-RL~\cite{frans2025diffusion,intelligence2025pi} to the aggregated dataset of original demonstrations, real rollouts, and imagined corrections.
We condition action generation on a binary advantage label \(y\in\{0,1\}\).
All samples from original demonstrations and successful real rollouts are assigned \(y=1\).
For failed real rollouts, we assign \(y=1\) before failure onset and \(y=0\) from failure onset onward.
Retained imagined corrections receive \(y=1\) according to the selection procedure in Section~\ref{method:correction}.

During training, we replace \(y\) with a null label \(\varnothing\) with probability \(p_{\mathrm{drop}}\), yielding \(\tilde{y}\).
The action expert receives the adaRMSNorm condition
\[
\tilde{c}_{\mathrm{adaRMS}}=c_\sigma+\lambda_f c_f+\lambda_y c_y(\tilde{y}),
\]
where \(c_\sigma\), \(c_f\), and \(c_y(\tilde{y})\) encode the denoising timestep, tactile history, and advantage label, respectively, and \(\lambda_f\) and \(\lambda_y\) weight their corresponding contributions.
Label dropout preserves the VLM prefix, tactile history, and denoising-timestep conditioning.

Given a target action chunk \(a_t\), Gaussian noise \(\epsilon\sim\mathcal{N}(0,I)\), and a sampled denoising timestep \(\sigma\in[0,1]\), we construct \(x_\sigma=\sigma\epsilon+(1-\sigma)a_t\) and minimize the flow-matching objective:
\[
\mathcal{L}_{\pi}=\mathbb{E}\!\left[\left\|u_\theta(x_\sigma,\sigma\mid z_t,\tilde{c}_{\mathrm{adaRMS}})-(\epsilon-a_t)\right\|_2^2\right].
\]

\noindent\textbf{Guided Inference.}
At inference, classifier-free guidance~\cite{ho2022classifier} combines the velocity predictions obtained with the positive and null labels:
\[
u_{\mathrm{CFG}}=u_{\varnothing}+w(u_1-u_{\varnothing}),
\]
where \(u_1\) and \(u_{\varnothing}\) share the same noisy action, denoising timestep, and observation conditions.
The guidance scale \(w\geq1\) controls the bias toward the positive-label action distribution.
We generate action chunks by integrating the guided velocity field from Gaussian noise at \(\sigma=1\) to \(\sigma=0\).
The updated policy \(\pi_\theta^{(k+1)}\) is then deployed to collect real rollouts for the next correction iteration.
We also update the visuo-tactile generation model and IDVM at each iteration, so that both components co-evolve with the robot policy throughout the correction loop.
\begin{table*}[t!]
    \caption{
        \textbf{Quantitative Results on Real-World Tasks.}
        We report task score (TS)  and completion steps (CS) across six contact-rich manipulation tasks.
        CS is averaged over successful episodes only.
        Iterations 1 and 2 denote the first and second rounds of post-training, respectively.
        Ave reports the mean TS and CS across all tasks.
    }
    \label{tab:real_world_results}
    \centering
    \begingroup
    \definecolor{TACOAccent}{RGB}{91,80,158}
    \definecolor{TACOShade}{RGB}{241,239,250}

    \includegraphics[width=0.96\textwidth]{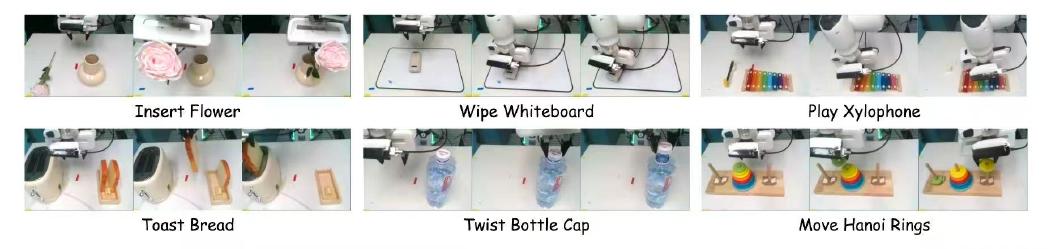}
    \par\smallskip

    \resizebox{0.96\textwidth}{!}{%
    \setlength{\tabcolsep}{5pt}
    \renewcommand{\arraystretch}{1.10}
    \begin{tabular}{l*{7}{cc}}
        \toprule
        \textbf{Method}
        & \multicolumn{2}{c}{\makecell{\textit{Insert}\\\textit{Flower}}}
        & \multicolumn{2}{c}{\makecell{\textit{Wipe}\\\textit{Whiteboard}}}
        & \multicolumn{2}{c}{\makecell{\textit{Twist}\\\textit{Bottle Cap}}}
        & \multicolumn{2}{c}{\makecell{\textit{Play}\\\textit{Xylophone}}}
        & \multicolumn{2}{c}{\makecell{\textit{Toast}\\\textit{Bread}}}
        & \multicolumn{2}{c}{\makecell{\textit{Move}\\\textit{Hanoi Rings}}}
        & \multicolumn{2}{c}{Ave} \\
        \cmidrule(lr){2-3}
        \cmidrule(lr){4-5}
        \cmidrule(lr){6-7}
        \cmidrule(lr){8-9}
        \cmidrule(lr){10-11}
        \cmidrule(lr){12-13}
        \cmidrule(lr){14-15}
        & TS & CS
        & TS & CS
        & TS & CS
        & TS & CS
        & TS & CS
        & TS & CS
        & TS & CS \\
        \midrule

        Base Policy
        & 0.500 & 250
        & 0.500 & 151
        & 0.425 & 131
        & 0.450 & 132
        & 0.300 & 183
        & 0.075 & 266
        & 0.375 & 185.5 \\

        \midrule
        \rowcolor{gray!15}
        \multicolumn{15}{c}{\textit{Iteration 1}} \\
        \midrule

        RFT
        & 0.550 & 274
        & 0.525 & 120
        & 0.500 & 62
        & 0.475 & 128
        & 0.300 & 189
        & 0.050 & 194
        & 0.400 & 161.2 \\

        DreamGen
        & 0.650 & 232
        & 0.550 & 106
        & 0.450 & 70
        & 0.600 & 130
        & 0.450 & 200
        & 0.150 & 140
        & 0.475 & 146.3 \\

        \rowcolor{TACOShade}
        \textcolor{TACOAccent}{\textbf{TACO}}
        & 0.700 & 207
        & 0.550 & 95
        & 0.850 & 56
        & 0.625 & 115
        & 0.675 & 184
        & 0.500 & 130
        & 0.650 & 131.2 \\

        \midrule
        \rowcolor{gray!15}
        \multicolumn{15}{c}{\textit{Iteration 2}} \\
        \midrule

        RFT
        & 0.525 & 289
        & 0.575 & 133
        & 0.475 & 79
        & 0.525 & 125
        & 0.350 & \textbf{177}
        & 0.100 & 184
        & 0.425 & 164.5 \\

        DreamGen
        & 0.650 & 210
        & 0.600 & 111
        & 0.550 & 67
        & \textbf{0.800} & 104
        & 0.500 & 185
        & 0.200 & \textbf{113}
        & 0.550 & 131.7 \\

        \rowcolor{TACOShade}
        \textcolor{TACOAccent}{\textbf{TACO}}
        & \textbf{0.925} & \textbf{169}
        & \textbf{0.675} & \textbf{87}
        & \textbf{0.975} & \textbf{52}
        & 0.775 & \textbf{97}
        & \textbf{0.825} & \textbf{177}
        & \textbf{0.775} & 120
        & \textbf{0.825} & \textbf{117.0} \\
        \bottomrule
    \end{tabular}%
    }
    \endgroup
\end{table*}

\section{Experiments}
% Section~\ref{exp:setup} introduces the experimental setup. Section~\ref{exp:main} reports results on real-world tasks through two iterations of post-training, while Section~\ref{exp:ablation} presents ablation studies of key components. Section~\ref{exp:analysis} analyzes action distributions and OOD adaptation.

\subsection{Experiment Setup}
\label{exp:setup}
\noindent\textbf{Data Collection.} We evaluate TACO on real-world contact-rich manipulation tasks using a single-arm Franka Research 3 robot, equipped with a front-view camera and two Xense tactile sensors mounted on the gripper. We perform six tasks: 1) \textit{Insert Flower}, 2) \textit{Wipe Whiteboard}, 3) \textit{Twist Bottle Cap}, 4) \textit{Play Xylophone}, 5) \textit{Toast Bread} and 6) \textit{Move Hanoi Rings}. Each task includes 50 demonstration trajectories collected via SpaceMouse teleoperation. 

\noindent\textbf{Baselines.}
All methods share a common initialization obtained by fine-tuning \(\pi_{0.5}\)~\cite{intelligence2025pi_} on 50 demonstrations per task.
\textit{Base Policy} evaluates this initialization without further post-training.
\textit{Rejection Sampling Fine-Tuning (RFT)} retains successful real rollouts at each iteration and combines them with the original demonstrations for supervised policy updates.
\textit{DreamGen}~\cite{jang2025dreamgen} generates robot videos conditioned on initial observations and task instructions, infers pseudo-action labels, and uses the resulting synthetic trajectories for policy post-training.

\noindent\textbf{Training and Evaluation Details.}
TACO performs two iterations per task.
At each iteration, we collect real rollouts with the current policy, generate imagined corrections with the compositional tactile world model, and apply CFG-RL post-training to the accumulated dataset of demonstrations, real rollouts, and imagined corrections.
Each method is evaluated over 40 independent episodes per task with randomized initial tabletop object positions.
We report task score (TS), which credits partial task completion, and average completion steps (CS) over successful episodes, with fewer steps indicating greater execution efficiency.

\begin{figure*}[t]
  \centering
  \includegraphics[width=\textwidth]{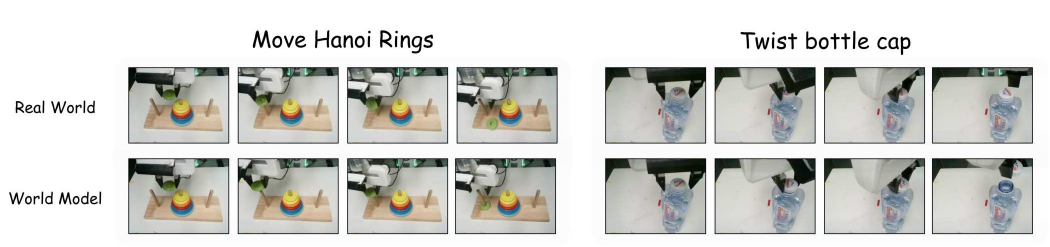}
  \caption{\textbf{Visualization of imagined correction data.} The compositional tactile world model generates locally consistent imagined corrections (bottom) that recover failed contact interactions (top).
  }
  \label{fig:correction_vis}
\end{figure*}

\begin{table*}[t]
    \caption{
        \textbf{Ablation of the Compositional Tactile World Model.}
        $V$, $F$, $A$, and $R$ denote video, tactile signals,
        actions, and value scores, respectively.
        SA, VOC, and FPL denote success AUROC, video-frame progress
        rank correlation, and failure point localization accuracy.
        Replay Pass and Policy TS are averaged across six tasks.
    }
    \label{tab:correction_ablation}
    \centering
    \begingroup
    \definecolor{TACOAccent}{RGB}{91,80,158}
    \definecolor{TACOShade}{RGB}{241,239,250}
    \footnotesize
    \setlength{\tabcolsep}{3pt}
    \renewcommand{\arraystretch}{1.2}

    \begin{tabularx}{\textwidth}{
        @{}
        >{\raggedright\arraybackslash}p{0.07\textwidth}
        *{11}{>{\centering\arraybackslash}X}
        @{}
    }
        \toprule
        \multirow{2}{*}{\textbf{Setting}}
        & \multicolumn{2}{c}{\textbf{Generation}}
        & \multicolumn{2}{c}{\textbf{IDVM}}
        & \multirow{2}{*}{\makecell{\textbf{Tactile}\\NRMSE $\downarrow$}}
        & \multirow{2}{*}{\makecell{\textbf{Action}\\Loss $\downarrow$}}
        & \multicolumn{3}{c}{\textbf{Value Evaluation}}
        & \multirow{2}{*}{\makecell{\textbf{Replay}\\Pass (\%) $\uparrow$}}
        & \multirow{2}{*}{\makecell{\textbf{Policy}\\TS $\uparrow$}} \\
        \cmidrule(lr){2-3}
        \cmidrule(lr){4-5}
        \cmidrule(lr){8-10}
        & Input & Output & Input & Output & &
        & SA $\uparrow$ & VOC $\uparrow$ & FPL $\uparrow$ & & \\
        \midrule

        (i)
        & $V$ & $V$ & $V$ & $A+R+F$
        & 0.42 & 0.025 & 0.87 & 0.78 & 0.87
        & 32.0 & 0.275 \\

        (ii)
        & $V+F$ & $V+F$ & $V$ & $A+R$
        & \textbf{0.28} & 0.038 & 0.91 & 0.88 & 0.90
        & 86.0 & 0.650 \\

        \rowcolor{TACOShade}
        \textcolor{TACOAccent}{\textbf{TACO}}
        & $V+F$ & $V+F$ & $V+F$ & $A+R$
        & \textbf{0.28} & \textbf{0.019}
        & \textbf{0.98} & \textbf{0.94} & \textbf{0.95}
        & \textbf{98.0} & \textbf{0.825} \\
        \bottomrule
    \end{tabularx}
    \endgroup
\end{table*}

\begin{table}[t]
    \caption{
        \textbf{Component Ablations.}
        Task score for ablations of correction and post-training components.
    }
    \label{tab:component_ablation}
    \centering
    \begingroup
    \definecolor{TACOAccent}{RGB}{91,80,158}
    \definecolor{TACOShade}{RGB}{241,239,250}
    \definecolor{AWRAccent}{RGB}{57,112,151}
    \definecolor{CFGRLAccent}{RGB}{173,70,105}
    \footnotesize
    \setlength{\tabcolsep}{3pt}
    \renewcommand{\arraystretch}{1.2}

    \begin{tabularx}{\linewidth}{
        @{}
        >{\raggedright\arraybackslash}p{0.50\linewidth}
        *{2}{>{\centering\arraybackslash}X}
        c
        @{}
    }
        \toprule
        \textbf{Setting}
        & \makecell{\textit{Insert}\\\textit{Flower}}
        & \makecell{\textit{Wipe}\\\textit{Whiteboard}}
        & \textbf{Ave} \\
        \midrule

        w/o knowledge insulation
        & 0.600 & 0.350 & 0.475 \\

        w/o failure-point anchoring
        & 0.800 & 0.250 & 0.525 \\

        w/o correction filtering
        & 0.850 & 0.500 & 0.675 \\

        \textbf{%
            \textcolor{TACOAccent}{TACO}%
            (\textcolor{AWRAccent}{AWR})%
        }
        & 0.875 & 0.625 & 0.750 \\

        \rowcolor{TACOShade}
        \textbf{%
            \textcolor{TACOAccent}{TACO}%
            (\textcolor{CFGRLAccent}{CFG-RL})%
        }
        & \textbf{0.925} & \textbf{0.675} & \textbf{0.800} \\
        \bottomrule
    \end{tabularx}
    \endgroup
\end{table}

\begin{table}[t]
    \caption{
        \textbf{Scaling of Imagined Correction Data.}
        Task score at different real-to-imagined data ratios.
    }
    \label{tab:correction_scaling}
    \centering
    \begingroup
    \definecolor{TACOAccent}{RGB}{91,80,158}
    \definecolor{TACOShade}{RGB}{241,239,250}
    \footnotesize
    \setlength{\tabcolsep}{3pt}
    \renewcommand{\arraystretch}{1.2}

    \begin{tabularx}{\linewidth}{
        @{}
        >{\raggedright\arraybackslash}p{0.50\linewidth}
        *{2}{>{\centering\arraybackslash}X}
        c
        @{}
    }
        \toprule
        \textbf{Setting}
        & \makecell{\textit{Insert}\\\textit{Flower}}
        & \makecell{\textit{Toast}\\\textit{Bread}}
        & \textbf{Ave} \\
        \midrule

        Base Policy
        & 0.500 & 0.300 & 0.400 \\

        Real:Imagined = $1{:}2$
        & 0.700 & 0.550 & 0.625 \\

        Real:Imagined = $1{:}4$
        & 0.875 & 0.775 & 0.825 \\

        Real:Imagined = $1{:}8$
        & 0.925 & 0.825 & 0.875 \\

        \rowcolor{TACOShade}
        Real:Imagined = $1{:}16$
        & \textbf{0.950} & \textbf{0.900} & \textbf{0.925} \\
        \bottomrule
    \end{tabularx}
    \endgroup
\end{table}

\subsection{Main Results}
\label{exp:main}
\noindent\textbf{Visualization of Imagined Corrections.}
As shown in Figure~\ref{fig:correction_vis}, the compositional tactile world model generates local visuo-tactile corrections (bottom) from failure-adjacent states in real rollouts (top). 
We take two representative tasks as examples.
In \textit{Move Hanoi Rings}, the real rollout misaligns the ring with the peg and fails to insert it, whereas the imagined sequence depicts a contact adjustment that realigns the ring and seats it onto the peg.
In \textit{Twist Bottle Cap}, the real rollout contacts the cap but slips without sustaining effective twisting torque, whereas the imagined sequence depicts sustained contact and completion of the twist.
The generated tactile trajectories complement visual motion with tactile information that is difficult to infer from vision alone.

\noindent\textbf{Quantitative and Qualitative Results.}
Table~\ref{tab:real_world_results} compares performance across six real-world tasks over two post-training iterations.
After the second iteration, TACO achieves an average task score of 0.825, compared with 0.375 for the base policy, 0.425 for \textit{RFT}, and 0.550 for \textit{DreamGen}.
The modest gains of \textit{RFT} are consistent with continued reinforcement of the narrow demonstration manifold: successful-rollout aggregation strengthens behaviors already supported by the policy without explicitly targeting corrections at failure-adjacent contact states.
\textit{DreamGen} expands supervision through generated visual trajectories, while TACO additionally models tactile trajectories and incorporates them into corrective action labeling.
This provides explicit contact information for learning recovery behaviors that are difficult to infer from RGB alone.
Qualitatively, the illustrated corrections preserve the task objective while modifying local contact strategies, including realignment before insertion and sustained contact during twisting.
Knowledge-insulated post-training keeps the pretrained backbone used for pre-contact perception and grounding fixed while adapting the action expert to these contact adjustments.
This combines the visual-language representations needed for target approach with corrective supervision for contact-sensitive manipulation.
After the second iteration, TACO also achieves lower CS than the base policy and the baseline methods on most tasks.
The lower CS suggests more efficient successful executions and is consistent with reduced pauses, redundant motions, and hesitation during contact transitions.

\begin{figure*}[t]
  \centering
  \includegraphics[width=\textwidth]{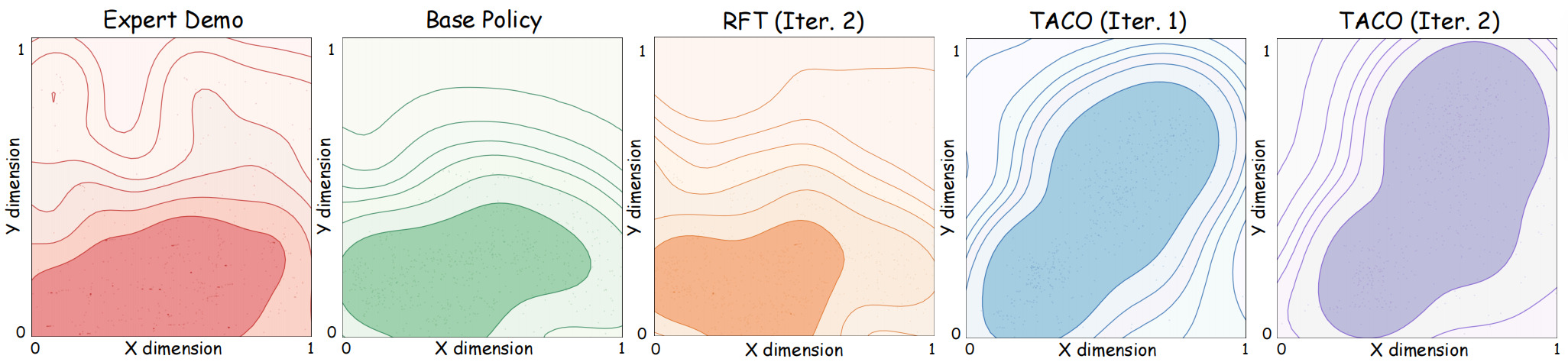}
  \caption{\textbf{Action Distribution Analysis.} We project the end-effector poses (X-Y dimensions) from 40 successful rollouts under different configurations in \textit{Insert Flower} task.
  }
  \label{fig:action_distribution}
\end{figure*}

\subsection{Ablation Study}
\label{exp:ablation}
\noindent\textbf{Compositional Tactile World Model.}
In Table~\ref{tab:correction_ablation}, \textit{Setting (i)} generates video only and uses the inverse dynamics and value model (IDVM) to infer actions, values, and tactile trajectories from it.
\textit{Setting (ii)} jointly generates video and tactile trajectories but feeds only video to the IDVM; \textit{TACO} shares this generator and adds tactile input.
On held-out real segments, Tactile NRMSE averages channel-wise RMSE normalized by fixed training-set standard deviations; Action Loss uses action-only Smooth~L1 with identical preprocessing and loss reduction.
For value evaluation, we collect additional successful and failed real-world rollouts across six tasks exclusively for testing, with synchronized visuo-tactile observations, outcome labels, and manually annotated failure points.
SA measures episode-level outcome AUROC on both subsets; VOC measures the rank correlation between predicted values and frame order in successful rollouts; FPL measures failure localization accuracy in failed rollouts.
Replay Pass reports the fraction of 50 randomly sampled filtered candidates per task and setting judged valid through real-robot execution and manual inspection.
Policy TS credits partial completion after two post-training iterations.
The value metrics, Replay Pass, and Policy TS are averaged equally across tasks.
\textit{Setting (i)} yields the lowest Replay Pass and Policy TS, while \textit{Setting (ii)} improves both with joint visuo-tactile generation.
With the generator unchanged, \textit{TACO} halves Action Loss relative to \textit{Setting (ii)} and improves all three value metrics, Replay Pass, and Policy TS.
These results suggest that tactile cues enrich visual imagination with contact information and improve action inference and value estimation, yielding more effective corrective supervision.

\noindent\textbf{Component Ablations.}
Table~\ref{tab:component_ablation} evaluates the correction and post-training components on \textit{Insert Flower} and \textit{Wipe Whiteboard}.
The substantial drop under \textit{w/o knowledge insulation} highlights the importance of preserving pretrained representations when incorporating tactile supervision. Failures in this setting often occur during the pre-contact phase, suggesting that uninsulated tactile adaptation can disrupt the pretrained visual-language representations needed for target perception and grounding.
The gains from imagined corrections also depend on where they start and how they are selected.
Uniform anchor sampling from failed rollouts in \textit{w/o failure-point anchoring} is particularly detrimental to \textit{Wipe Whiteboard}, emphasizing the value of targeting failure-adjacent states.
Even with these anchors retained, \textit{w/o correction filtering} lowers task score when validity checks and value-based selection are replaced by uniform sampling from unfiltered candidates.
Under the same correction pipeline, replacing advantage-weighted regression with CFG-RL improves both tasks and raises average task score from 0.750 to 0.800.

\noindent\textbf{Scaling of Imagined Correction Data.}
As shown in Table~\ref{tab:correction_scaling}, we further evaluate the effectiveness of imagined correction data on \textit{Insert Flower} and \textit{Toast Bread} under the same training recipe.
Compared with \textit{Base Policy}, incorporating imagined corrections consistently improves task score.
As the real-to-imagined data ratio changes from $1{:}2$ to $1{:}16$, task score increases from 0.700 to 0.950 on \textit{Insert Flower} and from 0.550 to 0.900 on \textit{Toast Bread}.
Notably, $1{:}16$ further outperforms $1{:}8$ on both tasks, suggesting that larger amounts of imagined corrections provide broader coverage of failure-adjacent contact states.
These results show that scaling imagined corrections improves policy post-training while reducing reliance on manually collected expert corrections.

\begin{table}[t]
  \caption{\textbf{OOD Adaptation.}
    Task score under unseen backgrounds (Bg.),
    objects (Obj.), and positions (Pos.).}
  \label{tab:generalization}
  \centering
  \begingroup
  \definecolor{TACOAccent}{RGB}{91,80,158}
  \definecolor{TACOShade}{RGB}{241,239,250}
  \footnotesize
  \setlength{\tabcolsep}{3pt}
  \renewcommand{\arraystretch}{1.22}

  \begin{tabularx}{\linewidth}{@{}l*{7}{>{\centering\arraybackslash}X}@{}}
    \toprule
    \multirow{2}{*}{\textbf{Method}}
    & \multicolumn{3}{c}{\textit{Insert Flower}}
    & \multicolumn{3}{c}{\textit{Wipe Whiteboard}}
    & \multirow{2}{*}{\textbf{Ave}} \\
    \cmidrule(lr){2-4}\cmidrule(lr){5-7}
    & Bg. & Obj. & Pos. & Bg. & Obj. & Pos. & \\
    \midrule

    Base Policy
    & 0.200 & 0.350 & 0.125
    & 0.250 & 0.300 & 0.125
    & 0.225 \\

    \rowcolor{TACOShade}
    \textcolor{TACOAccent}{\textbf{TACO}}
    & \textbf{0.800} & \textbf{0.850} & \textbf{0.550}
    & \textbf{0.750} & \textbf{0.825} & \textbf{0.425}
    & \textbf{0.700} \\
    \bottomrule
  \end{tabularx}
  \endgroup
\end{table}

\subsection{Analysis}
\label{exp:analysis}
\noindent\textbf{Action Distribution Analysis.} 
To further analyze how TACO improves action learning, we visualize the action distributions on the \textit{Insert Flower} task. During evaluation, the initial positions of the flower and vase are fixed.  Following the task setup, we project the end-effector (EE) poses along the world-frame $z$-axis and compare their distributions in the $x$-$y$ plane. For each policy, we collect 40 successful rollouts and record the EE poses. We compare five configurations: (a) expert demonstrations, (b) base policy, (c) RFT (Iter.2), (d) TACO (Iter.1), and (e) TACO (Iter.2). 
As shown in Figure~\ref{fig:action_distribution}, the base policy (b) exhibits a narrowly concentrated action distribution around the demonstration manifold, making it sensitive to accumulated execution errors. \textit{RFT} reinforces the trajectories the base policy already covers and therefore cannot expand beyond the original manifold, whereas our method progressively yields a broader distribution across iterations (d, e), exposing the policy to more
diverse successful adjustment behaviors. These results suggest that TACO broadens the action space of the policy beyond the narrow demonstration manifold, enabling it to recover and complete the task even in scenarios unseen in the expert data.

\noindent\textbf{Adaptation to Unseen Scenarios.}
We evaluate fast adaptation to unseen backgrounds, unseen objects, and unseen positions on \textit{Insert Flower} and \textit{Wipe Whiteboard}.
We compare the base policy, fine-tuned only on in-domain expert demonstrations, with the policy obtained after one TACO iteration using imagined corrections generated from out-of-domain (OOD) rollout states.
Table~\ref{tab:generalization} shows that the base policy degrades across all three shifts, whereas TACO improves task score in every setting, increasing the average from 0.225 to 0.700.
By generating imagined corrections around OOD states, TACO exposes the policy to diverse recovery behaviors without requiring additional expert demonstrations in the target scenarios.
Background changes primarily challenge visual grounding, while novel objects and placements can also alter the approach motions and contact conditions required for task completion.
Across these settings, changes in visual observations, tactile interactions, and required actions extend beyond the world model's training distribution.
The improvements suggest that the compositional tactile world model can provide effective corrective supervision under these shifts.
Together with the in-domain results, these findings support TACO's ability to improve contact-rich execution and efficiently adapt robot policies to novel objects, placements, and visual environments.
\section{Conclusion}
\label{sec:conclusion}
We presented TACO, a framework for scalable robot policy post-training in contact-rich manipulation built on a compositional tactile world model.
Through a \textit{Recognize--Imagine--Label} loop, TACO generates and labels local visuo-tactile corrections from failure-adjacent states, then selects candidates through validity filtering and progress-based ranking.
The selected corrections are iteratively aggregated with demonstrations and real rollouts for post-training, combining knowledge-insulated tactile adaptation with CFG-RL using binary advantage labels.
Across six real-world tasks, TACO improves the average task score from 0.375 to 0.825 after two post-training iterations.
Further analyses show that TACO broadens the policy's action distribution across iterations and effectively adapts to unseen scenarios without additional expert demonstrations.
\FloatBarrier
% Native ieeeconf hook to balance this working draft's final reference page.
% Revisit this reference number after substantial content edits.
\bibliographystyle{IEEEtran}
\bibliography{main}
% Append after the references without leaving a nearly empty reference page.
\par\bigskip
% Appendix revised from the supplied text; main-text sources are unchanged.
\raggedbottom
\makeatletter
\setlength{\@fptop}{0pt}
\setlength{\@fpsep}{16pt plus 2pt minus 2pt}
\setlength{\@fpbot}{0pt plus 1fil}
\setlength{\@dblfptop}{0pt}
\setlength{\@dblfpsep}{16pt plus 2pt minus 2pt}
\setlength{\@dblfpbot}{0pt plus 1fil}
\makeatother

% Muted colors for the appendix algorithm; no change to main-text styling.
\definecolor{TACOPseudoInk}{HTML}{443B78}
\definecolor{TACOPseudoTint}{HTML}{F1EFF9}
\definecolor{TACOPseudoTeal}{HTML}{286F68}
\definecolor{TACOPseudoBlue}{HTML}{3A5D8F}
\definecolor{TACOPseudoGold}{HTML}{91631D}
\definecolor{TACOPseudoMuted}{HTML}{7B8293}
\makeatletter
\newcommand{\floatc@tacopseudocode}[2]{%
  \colorbox{TACOPseudoTint}{%
    \parbox{\dimexpr\linewidth-2\fboxsep\relax}{%
      \small\strut\textcolor{TACOPseudoInk}{\textbf{#1}\quad#2}\strut}}\par}
\newcommand{\tacocodecaption}[1]{%
  \refstepcounter{algorithm}%
  \addcontentsline{loa}{algorithm}{\protect\numberline{\thealgorithm}#1}%
  \floatc@tacopseudocode{Algorithm~\thealgorithm}{#1}\vspace{5pt}}
\newcommand{\fs@tacopseudocode}{%
  \def\@fs@cfont{\bfseries}\let\@fs@capt\floatc@tacopseudocode
  \def\@fs@pre{{\color{TACOPseudoInk}\hrule height0.6pt}\kern2pt}%
  \def\@fs@mid{\kern5pt}%
  \def\@fs@post{\kern4pt{\color{TACOPseudoMuted!45}\hrule height0.4pt}}%
  \let\@fs@iftopcapt\iftrue}
\makeatother
\floatstyle{tacopseudocode}
\restylefloat{algorithm}
\newcommand{\tacophase}[2]{%
  \Statex\textcolor{#1}{\rule[-0.2ex]{1.8pt}{1.6ex}\hspace{0.45em}\textbf{#2}}}

\useRomanappendicesfalse
\appendices

\section{Real-World Setup}
\label{apsec:A}

\subsection{Real-World Robot Configuration}
\begin{figure}[H]
  \centering
  \includegraphics[width=\linewidth]{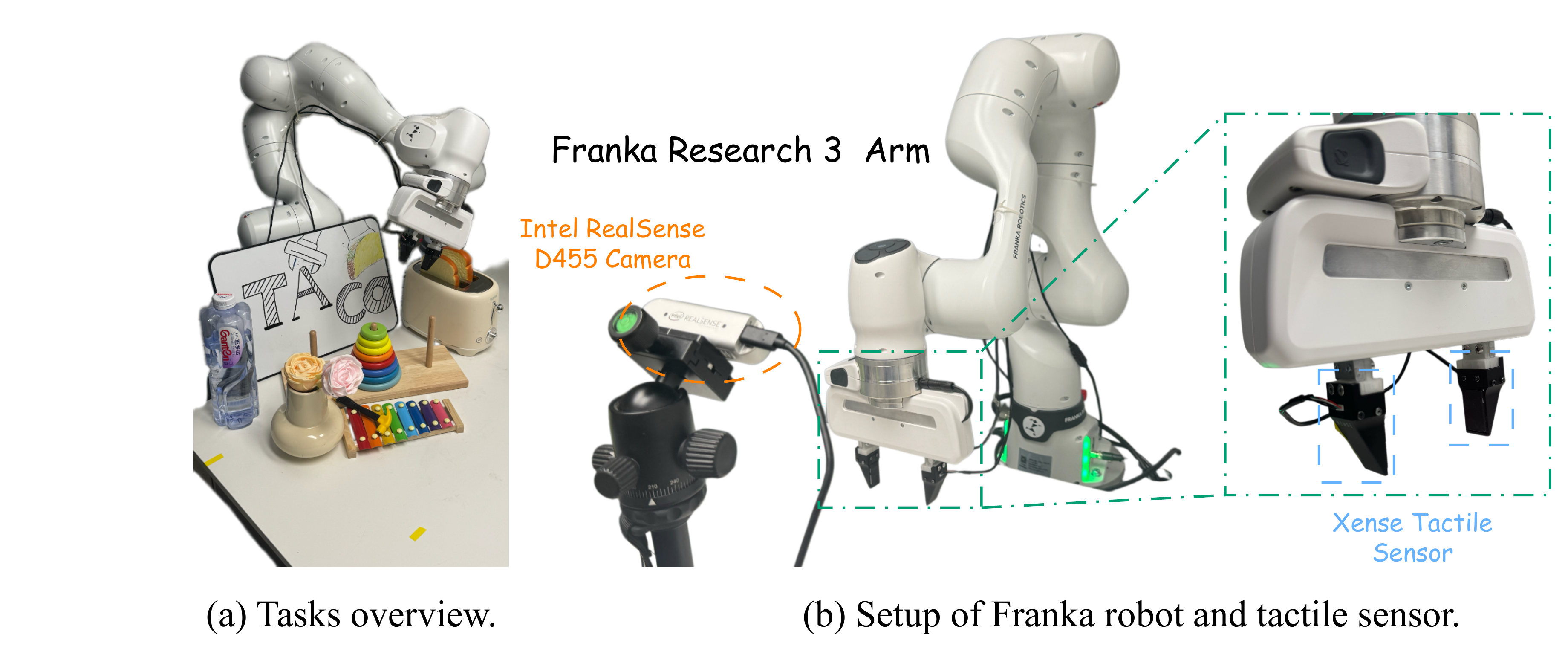}
  \caption{\textbf{Real-world robot setup and experimental assets.} We use a single-arm Franka Research 3 (FR3) platform equipped with a parallel-jaw gripper and fingertip-mounted Xense tactile sensors for 6D force/torque sensing. A front-view Intel RealSense D455 camera provides global RGB observations of the workspace, together with the object assets used across our real-world contact-rich manipulation tasks.}
  \label{fig:setup}
\end{figure}

As shown in Figure~\ref{fig:setup}, our single-arm platform is built on a 7-DoF Franka Research 3 (FR3) manipulator equipped with a parallel-jaw gripper whose fingertips integrate Xense tactile sensors, enabling 6D force/torque sensing for contact-rich precision manipulation tasks. 
The perception system employs a single front-view Intel RealSense D455 camera, which provides a global view of the workspace using RGB images at a resolution of $640 \times 480$ pixels.
At each timestep, the policy observation consists of the RGB image, the 6D force/torque readings from the Xense tactile sensors, and the robot's proprioceptive state.

\begin{figure*}[!t]
  \centering
  \includegraphics[width=0.78\textwidth]{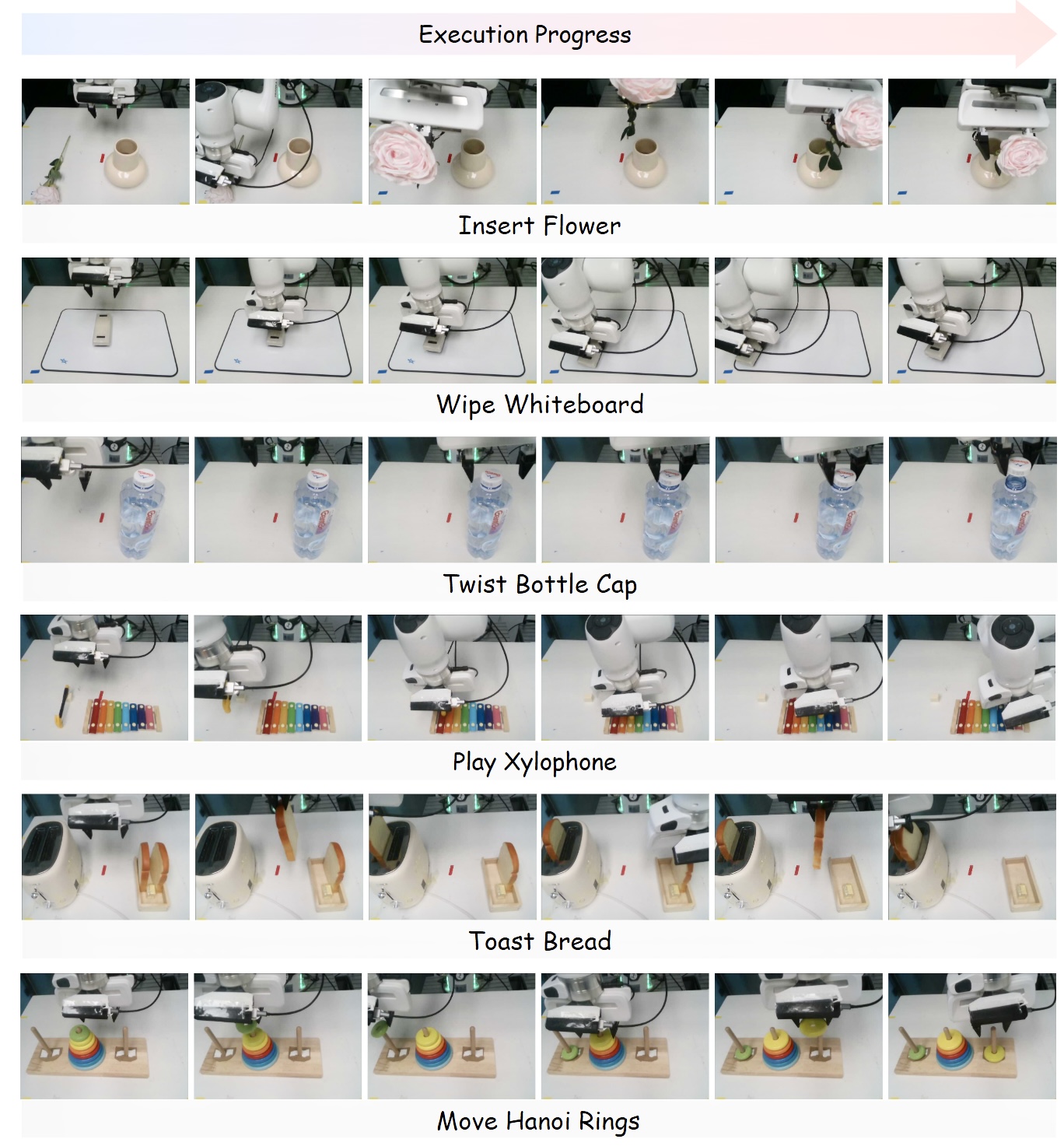}
\caption{
    \textbf{Robot execution progress in real-world tasks.} We visualize key frames of the robot’s execution process from the front camera view in real-world tasks.
}
  \label{fig:task_visualization2}
\end{figure*}

\subsection{Control and Data Collection}
The end-effector pose, gripper state, 6D force/torque readings from the Xense tactile sensors, and other proprioceptive signals are synchronized with camera observations through a real-time communication interface. All demonstrations are collected via teleoperation using a SpaceMouse 3D controller, with the operator receiving real-time feedback from the camera view. As shown in Figure~\ref{fig:task_visualization2}, we visualize the execution progress for the six real-world tasks from the front camera view. Detailed stage decomposition and success criteria are as follows:
\begin{description}[leftmargin=0pt,labelindent=0pt,itemindent=0pt,labelsep=0.5em,style=unboxed]
    \item[\textit{1. Insert Flower.}]
The robot picks up a flower from a random location on the table and inserts it into a vase. 
S1: insert the flower into the vase. 
S1 is considered successful if the robot securely inserts the flower into the vase without dropping it or knocking over the vase.

    \item[\textit{2. Wipe Whiteboard.}]
The robot picks up an eraser and wipes off a star drawn at a random location on a whiteboard. 
S1: grasp the eraser; S2: wipe off the star. 
S1 is considered successful if the robot securely grasps and lifts the eraser. 
S2 is considered successful if the robot completely erases the star from the whiteboard.

    \item[\textit{3. Twist Bottle Cap.}]
The robot grasps the cap of a water bottle, applies a twisting motion to loosen it, and then lifts the cap away from the bottle. 
S1: grasp the bottle cap; S2: twist open and lift the cap. 
S1 is considered successful if the robot securely grips the cap with stable contact and without displacing the bottle. 
S2 is considered successful if the robot rotates the cap open and lifts it upward, fully separating it from the bottle without dropping the cap.

    \item[\textit{4. Play Xylophone.}] 
The robot picks up a mallet and sequentially strikes the 1st, 3rd, 5th, and 8th keys of an eight-key xylophone in order. 
S1: strike the 1st key; S2: strike the 3rd key; S3: strike the 5th key; S4: strike the 8th key. 
Each stage is considered successful if the robot accurately strikes the designated key with the mallet tip, while following the specified order and avoiding unintended strikes on adjacent keys.

    \item[\textit{5. Toast Bread.}] 
The robot sequentially picks up two slices of bread from a box and inserts them into a toaster. 
S1: grasp the first slice; S2: insert the first slice into the toaster; 
S3: grasp the second slice; S4: insert the second slice. 
Each stage is considered successful if the corresponding action is completed without dropping the bread or misaligning the insertion.

    \item[\textit{6. Move Hanoi Rings.}] 
The robot moves the top ring from the middle peg of a Hanoi tower (where all rings are initially stacked in size order) to the left empty peg, then moves the next ring to the right empty peg.
S1: grasp the top ring; S2: place the top ring on the left peg; 
S3: grasp the second ring; S4: place the second ring on the right peg. 
S1 is considered successful if the robot securely grasps and lifts the top ring. 
S2 is considered successful if the robot places the top ring on the left peg without dropping it. 
S3 is considered successful if the robot securely grasps and lifts the second ring. 
S4 is considered successful if the robot places the second ring on the right peg without dropping it or disturbing the first ring.
\end{description}
\subsection{Out-of-Domain (OOD) Scenario Construction}
\textbf{Unseen Background.} To evaluate robustness against visual context shifts, we construct background
OOD scenes by altering the tabletop appearance and ambient lighting. Specifically, a green tablecloth
with white stripes is spread over the workspace, and a red-green-blue disco light ball is introduced
to illuminate the tabletop, producing dynamically varying color reflections and highlights not
encountered during training. This setting tests the model's ability to maintain reliable perception and
stable manipulation under unfamiliar and visually distracting background conditions.

\textbf{Unseen Object.} To evaluate robustness to object-level semantic, geometric, and tactile variations,
target objects are replaced with substitutes that differ significantly from the training objects in both
visual appearance and physical properties. For example, the standard whiteboard eraser is replaced
with a sponge, which differs not only in color and texture but also in compliance and friction
characteristics, requiring the model to adapt both its visual recognition and its grasp force and
interaction strategy. This setting evaluates the model's ability to generalize to novel objects when
both perception and tactile dynamics deviate from training distributions.

\textbf{Unseen Position.} To evaluate spatial generalization, target items are placed in workspace regions
outside the spatial coverage of training demonstration trajectories. For instance, the star on the
whiteboard is drawn at a rear position far from the locations seen during data collection. This
setting evaluates the model's capability to handle unseen spatial coordinates while maintaining
accurate reaching and manipulation.

\section{Additional Method Details}
\label{apsec:B}
\subsection{Visuo-Tactile Generation Model Details}
In our implementation, we use correction segments of $T=49$ frames, including the anchor frame, and align the force sequence at the same temporal resolution. The video latents have $f=1+(T-1)/4=13$ temporal positions; $N_v$ counts all spatiotemporal video tokens. The DiT hidden dimension is $d=3072$.

After DiT denoising, the output is split back into video and force tokens. The video tokens are decoded by the original Wan unpatchify pathway, while the force-token output $Y^f\in\mathbb{R}^{B\times T\times d}$ is projected by a tactile head as
\[
\hat{u}^f=H_\eta(Y^f)\in\mathbb{R}^{B\times T\times12}.
\]

\subsection{Inverse Dynamics and Value Model Details}
The visual pathway uses a DINOv2-with-Registers backbone to extract \(37\times 37\) patch features with a feature dimension of 768. These features are further processed by a direction-aware decoder composed of four dilated convolution branches and angle-sensitive pooling, producing a 1024-dimensional visual embedding. The tactile pathway encodes the normalized 12-dimensional force-torque signal using a two-layer MLP with a hidden dimension of 128 and an output dimension of 256, yielding a 256-dimensional tactile embedding. The visual and tactile embeddings are concatenated into a shared 1280-dimensional representation. The action head contains a 512-dimensional hidden layer and predicts a 7-DoF end-effector action, while the progress head contains a 256-dimensional hidden layer and estimates signed task progress $\hat p_t\in\mathbb{R}$, with targets defined in Section~\ref{method:worldmodel}.

\subsection{Training Recipe Details}
\label{sec:appendix_training_recipe}

\noindent\textbf{Policy configuration.}
We use the publicly released $\pi_{0.5}$-DROID checkpoint, with a PaliGemma ($2$B) vision-language backbone and a $300$M action expert. The action expert predicts a $7$-DoF end-effector action chunk of horizon $30$, zero-padded to an action dimension of $32$.

\noindent\textbf{Tactile adaptation.}
The tactile encoder takes a history of $8$ synchronized left/right force--torque readings, with $12$ dimensions per timestep. It uses LayerNorm $\rightarrow$ Linear $\rightarrow$ SiLU $\rightarrow$ Linear $\rightarrow$ LayerNorm with a hidden dimension of $256$, producing the tactile conditioning vector $c_f$ with the same width as the action expert.

\noindent\textbf{CFG-RL implementation.}
An analogous label encoder with a hidden dimension of $256$ maps the CFG-RL label $\tilde y\in\{0,1,\varnothing\}$ to the conditioning vector $c_y(\tilde y)$. We set the label-dropout probability in Section~\ref{method:training} to $p_{\mathrm{drop}}=0.1$.

\noindent\textbf{Optimization.}
All policies are trained for $30{,}000$ steps with a batch size of $32$ using AdamW (gradient norm clipped to $1.0$) and an exponential moving average of $0.999$. The learning rate follows a cosine schedule with $300$ warmup steps and a peak of $5\times10^{-5}$. Training uses fully-sharded data parallelism across $8$ GPUs.

\section{Algorithm Pseudocode}
\label{apsec:C}
\label{sec:algorithm}
Algorithm~\ref{alg:taco} summarizes the complete post-training procedure.
\begin{algorithm}[!t]
\small
\tacocodecaption{TACO: Robot Policy Post-Training}
\label{alg:taco}
\algrenewcommand{\alglinenumber}[1]{\scriptsize\textcolor{TACOPseudoMuted}{#1:}}
\algrenewcommand{\algorithmicrequire}{\textcolor{TACOPseudoInk}{\textbf{Require:}}}
\algrenewcommand{\algorithmicensure}{\textcolor{TACOPseudoInk}{\textbf{Ensure:}}}
\algrenewcommand{\algorithmicfor}{\textcolor{TACOPseudoInk}{\textbf{for}}}
\algrenewcommand{\algorithmicdo}{\textcolor{TACOPseudoInk}{\textbf{do}}}
\algrenewcommand{\algorithmicif}{\textcolor{TACOPseudoInk}{\textbf{if}}}
\algrenewcommand{\algorithmicthen}{\textcolor{TACOPseudoInk}{\textbf{then}}}
\algrenewcommand{\algorithmicelse}{\textcolor{TACOPseudoInk}{\textbf{else}}}
\algrenewcommand{\algorithmicend}{\textcolor{TACOPseudoInk}{\textbf{end}}}
\algrenewcommand{\algorithmicreturn}{\textcolor{TACOPseudoInk}{\textbf{return}}}
\begin{algorithmic}[1]
\Require Policy $\pi_\theta$, generator $G_\psi$, IDVM $U_\phi$;
\Statex demonstrations $\mathcal D_{\rm demo}$; iterations $K_{\rm iter}$;
\Statex $T=49$, $H=T-1$, $K_1=32$, $K_2=8$;
\Statex progress window $\Delta$ and thresholds $\epsilon,\tau_p$
\Ensure Post-trained robot policy $\pi_\theta^*$
\State Warm-start $\pi_\theta$ on $\mathcal D_{\rm demo}$; set $y=1$
\State $\mathcal D_{\rm train}\gets\mathcal D_{\rm demo}$
\For{$k=1$ \textbf{to} $K_{\rm iter}$}
  \tacophase{TACOPseudoTeal}{1. Collect and recognize}
  \State Deploy $\pi_\theta$ to collect $\mathcal D_{\rm roll}^{(k)}$
  \State $\mathcal S\gets\varnothing$; $\mathcal D_{\rm corr}^{(k)}\gets\varnothing$
  \For{each rollout $\tau\in\mathcal D_{\rm roll}^{(k)}$}
    \State Predict signed progress $\{\hat p_t\}$ with $U_\phi$
    \If{$\operatorname{Succ}(\tau)=1$}
      \State Set $y_t\gets1$ for all $t$
    \Else
      \State Find nonterminal anchors $\mathcal A_\tau$ with
      \Statex \hspace{\algorithmicindent}$\hat p_{t+\Delta}-\hat p_t<\epsilon$, $t+\Delta\leq|\tau|$
      \If{$\mathcal A_\tau\neq\varnothing$}
        \State $t^*\gets$ earliest time in $\mathcal A_\tau$
        \State Set $y_t\gets1$ before $t^*$ and $0$ otherwise
        \State Add up to $10$ anchors from $\mathcal A_\tau$ to $\mathcal S$
      \Else
        \State Set $y_t\gets0$ for all $t$
      \EndIf
    \EndIf
    \State Add labeled rollout $\tau$ to $\mathcal D_{\rm train}$
  \EndFor
  \tacophase{TACOPseudoBlue}{2. Imagine and label}
  \For{each anchor $(\tau,t)\in\mathcal S$}
    \State Sample $K_1$ segments of $T$ frames from
    \Statex \hspace{\algorithmicindent}$G_\psi(\cdot\mid I_t,F_t,l)$ with independent seeds
    \State Label actions and progress with $U_\phi$
    \tacophase{TACOPseudoGold}{3. Filter and select}
    \State Discard segments failing kinematic or tactile checks
    \State Compute $g_j=\hat p_{t+H}^{(j)}-\hat p_t$
    \State Keep up to $K_2$ highest-gain valid segments
    \Statex \hspace{\algorithmicindent}with $g_j\geq\tau_p$; set their labels to $y=1$
    \State Add retained segments to $\mathcal D_{\rm corr}^{(k)}$
  \EndFor
  \tacophase{TACOPseudoInk}{4. CFG-RL policy update}
  \State $\mathcal D_{\rm train}\gets\mathcal D_{\rm train}\cup\mathcal D_{\rm corr}^{(k)}$
  \For{each minibatch from $\mathcal D_{\rm train}$}
    \State Compute the frozen VLM prefix $z_t$
    \State Replace $y$ by $\varnothing$ with probability $0.1$
    \State Encode tactile history and the resulting label
    \State Minimize $\mathcal L_\pi$ with CFG-RL label dropout
    \Statex \hspace{\algorithmicindent}as specified in Section~\ref{method:training}
  \EndFor
  \State Update $G_\psi$ and $U_\phi$ for the next iteration
\EndFor
\State \Return $\pi_\theta^*\gets\pi_\theta$
\end{algorithmic}
\end{algorithm}

\begin{figure*}[!t]
    \centering
    \includegraphics[width=\textwidth]{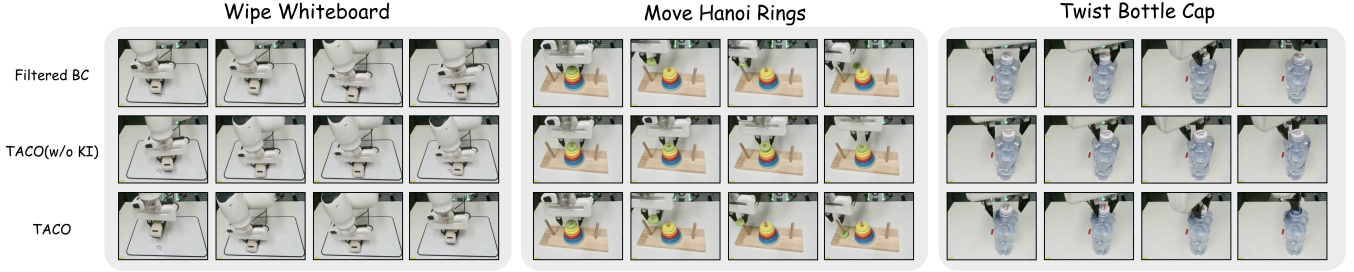}
    \caption{\textbf{Failure case analysis.} Representative rollouts of RFT, TACO~(w/o~KI), and TACO on \textit{Wipe Whiteboard}, \textit{Move Hanoi Rings}, and \textit{Twist Bottle Cap}. The two baselines fail at contact transitions in complementary ways, whereas TACO completes all three tasks. The original figure label ``Filtered BC'' denotes RFT.}
    \label{fig:failure_case}
\end{figure*}

\begin{figure*}[!t]
  \centering
  \includegraphics[width=\textwidth]{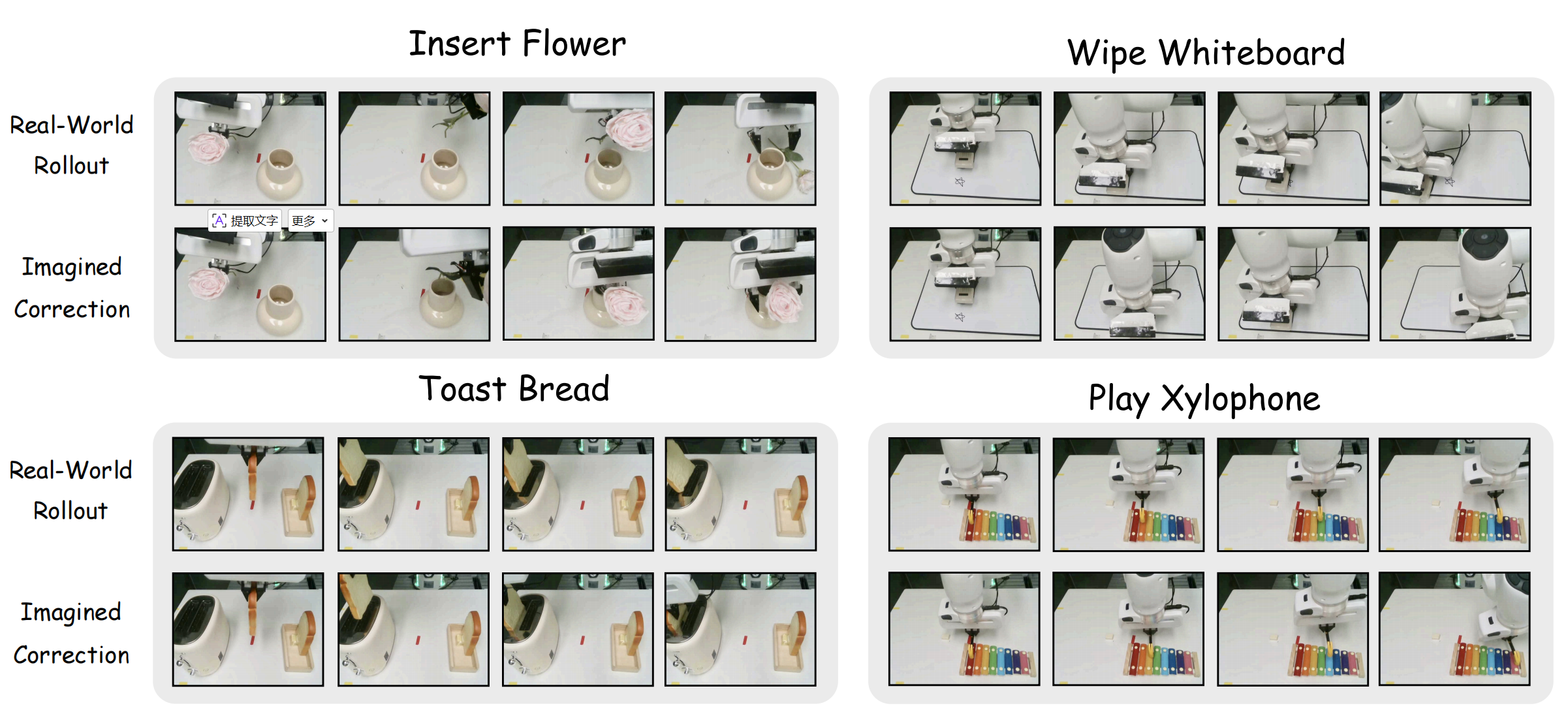}
  \caption{\textbf{Visualization of imagined correction data.} The compositional tactile world model generates locally consistent imagined corrections (bottom) that recover failed contact interactions (top).
  }
  \label{fig:correction_vis_appendix}
\end{figure*}

\section{Evaluation Details}
\label{apsec:D}

\subsection{Evaluation Metric Details}
\label{sec:appendix_generation}
We provide additional details of the metrics used in Section~\ref{exp:ablation}.

\noindent\textbf{Action Loss.} We report the validation loss of the inverse dynamics and value model on held-out real segments using action-only Smooth~L1 with identical preprocessing and loss reduction, measuring how well the model predicts actions from the visual and tactile signals. Lower values indicate more accurate action labeling.

\noindent\textbf{Tactile NRMSE.} To evaluate the tactile quality produced by the visuo-tactile generation model and consumed by the inverse dynamics and value model, we construct a held-out test set of real contact-rich segments with ground-truth 12-D force-torque readings. We report the mean channel-wise RMSE normalized by fixed training-set standard deviations, so that Tactile NRMSE reflects how faithfully the predicted tactile signals match real contact dynamics. Tactile predictions come from the IDVM in Setting~(i) and the joint visuo-tactile generator in Setting~(ii) and TACO.

\noindent\textbf{Success AUROC (SA).} We evaluate whether the model assigns higher trajectory-level scores to successful rollouts than to failed ones. Within each task, every held-out successful trajectory is paired with every held-out failed trajectory, yielding $N_+N_-$ pairs for $N_+$ successful and $N_-$ failed trajectories. A pair receives a score of $1$ if the successful trajectory has the higher predicted score, $0$ if it has the lower score, and $0.5$ for a tie. SA is the mean over all such pairs, equivalent to the empirical area under the ROC curve; higher values indicate better discrimination between successful and failed executions.

\noindent\textbf{Video-frame progress rank correlation (VOC).} We measure whether the progress estimates produced by the inverse dynamics and value model are correctly ordered along the true temporal progression of a trajectory. For each of the six tasks, we hold out a set of successful trajectories, shuffle their frames, and compute the rank correlation between the predicted progress and the ground-truth chronological order. Given two frames $t_A < t_B$, a correct prediction requires the estimated progress at $t_A$ to be lower than at $t_B$. The VOC score lies in $[-1, 1]$, with higher values indicating a more faithful understanding of task progression.

\noindent\textbf{Failure point localization accuracy (FPL).} To verify the Recognize step directly, we manually annotate failure-adjacent frames in held-out failed rollouts where contact begins to break down. We then check whether the inverse dynamics and value model, by detecting where progress stalls or decreases, flags these annotated frames as failure-adjacent. FPL reports the accuracy of this localization across the six tasks, so a higher FPL means the Recognize step more reliably identifies the contact transitions that need correction.

\noindent\textbf{Replay Pass.} Replay Pass reports the fraction of 50 randomly sampled filtered candidates per task and setting judged valid through real-robot execution and manual inspection.

\section{Additional Analysis}
\label{apsec:E}
\subsection{Failure Case Analysis}
\label{sec:appendix_failure}
Figure~\ref{fig:failure_case} shows representative rollouts of RFT, TACO~(w/o~KI), and TACO on three contact-rich tasks. The two baselines fail in distinct but complementary ways, and comparing them illustrates why both imagined visuo-tactile corrections and knowledge-insulated tactile adaptation are necessary.

\noindent\textbf{RFT: stalls at contact transitions.}
Trained only on its own successful rollouts, RFT inherits the narrow action distribution of the base policy and provides no recovery behavior at failure-adjacent contact states. As a result, it approaches the target but
stalls once contact becomes critical: in \textit{Wipe Whiteboard} it presses the eraser onto the board without applying enough force to remove the mark, in \textit{Move Hanoi Rings} it hovers around the peg but fails to align and seat the
ring, and in \textit{Twist Bottle Cap} it grips the cap without generating effective twisting torque. Because these contact-transition failures rarely self-correct within a rollout, they never enter the filtered successful set, so RFT keeps reproducing the same incomplete behavior.

\noindent\textbf{TACO~(w/o~KI): degraded pre-contact perception.}
TACO~(w/o~KI) learns from the same imagined corrections as TACO but fine-tunes the full VLA end-to-end, allowing tactile-action gradients to erode the pretrained visual-language priors. Without these priors, pre-contact perception
and spatial grounding degrade: the policy reaches the contact stage but with less accurate approach and alignment, leading to imprecise contact (e.g., off-target wiping, misaligned ring insertion, or unstable cap engagement). It
thus recovers some contact behaviors from the tactile corrections but loses the visual grounding needed to position itself reliably before contact.

\noindent\textbf{TACO: reliable approach and contact recovery.}
TACO combines imagined visuo-tactile corrections with knowledge-insulated tactile adaptation, preserving the pretrained visual-language priors for accurate pre-contact approach while learning failure-adjacent recovery from the tactile
corrections. It therefore both reaches the target reliably and adjusts contact force effectively, completing all three tasks: it wipes the mark clean, seats the ring onto the peg, and twists the cap open. These cases show that the two
components address complementary failure modes, and only their combination yields robust contact-rich execution.

\section{Additional Visualization}
\label{apsec:F}
\subsection{Additional Imagined Corrections Visualizations}
\label{sec:additional_correction_vis}
Beyond the two representative tasks shown in the main text, we further visualize imagined corrections for the remaining four contact-rich tasks in Figure~\ref{fig:correction_vis_appendix}, again pairing real-world failure rollouts (top) with the imagined corrections produced by the compositional tactile world model (bottom). Across all tasks, the world model localizes the failure-adjacent contact transition and imagines a segment that restores the intended contact behavior toward task completion.
In \textit{Insert Flower}, the real rollout brings the flower toward the vase but misaligns the stem with the narrow opening and fails to seat it, whereas the imagined correction realigns the approach and guides the stem into the vase. In \textit{Wipe Whiteboard}, the real rollout places the eraser over the target mark but applies insufficient contact force to remove it, whereas the imagined correction increases downward pressure and wipes the mark clean. In \textit{Play Xylophone}, the real rollout swings toward the target bar but under-contacts the strike, whereas the imagined correction adjusts the contact so the mallet lands firmly on the intended bar. In \textit{Toast Bread}, the real rollout grasps the bread but misaligns its placement into the slot, whereas the imagined correction corrects the alignment and seats the slice in place.
Consistent with the main-text examples, these corrections recover contact transitions that are weakly observable from RGB alone, yielding the corrective supervision used for post-training.

\end{document}